# Contrast Enhancement Estimation for Digital Image Forensics

LONGYIN WEN, GE Global Research Center
HONGGANG QI, University of Chinese Academy of Sciences
SIWEI LYU, University at Albany, State University of New York

Inconsistency in contrast enhancement can be used to expose imagel forgeries. In this work, we describe a new method to estimate contrast enhancement from a single image. Our method takes advantage of the nature of contrast enhancement as a mapping between pixel values, and the distinct characteristics it introduces to the image pixel histogram. Our method recovers the original pixel histogram and the contrast enhancement simultaneously from a single image with an iterative algorithm. Unlike previous methods, our method is robust in the presence of additive noise perturbations that are used to hide the traces of contrast enhancement. Furthermore, we also develop an effective method to to detect image regions undergone contrast enhancement transformations that are different from the rest of the image, and use this method to detect composite images. We perform extensive experimental evaluations to demonstrate the efficacy and efficiency of our method method.

## 1 INTRODUCTION

The integrity of digital images has been challenged by the development of sophisticated image editing tools (e.g., Adobe Photoshop), which can modify contents of digital images with minimal visible traces. Accordingly, the research field of digital image forensics [1] has experienced rapid developments in the past decade. Important cues to authenticate digital images and detect tampering can be found from various steps in the imaging pipeline, and one of such operations is contrast enhancement. Contrast enhancement is a nonlinear function of pixel values that changes the overall distribution of the pixel intensities, which are frequently used to enhance details of over-or under-exposed image regions. Commonly used contrast enhancement transforms include gamma correction, sigmoid stretching and histogram equalization[2]. In the relevance of forensics, recovering contrast enhancement is useful to reconstruct the processing history of an image. Also, detecting regions undergone different contrast enhancement can be used to expose a composite image.

There have been several methods to estimate contrast enhancement from an image [3–9]. However, these methods have two main limitations. First, most of these algorithms are designed for a specific type of contrast enhancement transform (*e.g.*, gamma correction). Second, these algorithms in general lack robustness with regards to noise perturbations that are added to hide the traces of contrast enhancement[10, 11].

In this work, we describe a general and efficient method to recover contrast enhancement from a single image. Our method exploits the fact that (i) contrast enhancement is a linear transformation of pixel histogram, (ii) pixel histogram after contrast enhancement tends to have more empty bins, and (iii) the effect of additive noise corresponds to convolution of pixel histogram. As a result, we formulate the estimation of contrast enhancement as an optimization problem seeking recovered pixel histogram to be consistent with the observed pixel histogram after contrast enhancement transform is applied, while with minimum number of empty bins. The original problem is intractable, so we further provide a continuous relaxation of it and an efficient numerical algorithm. Our formulation can handle the estimation of parametric and nonparametric contrast enhancement transforms alike, and is robust to additive noise perturbations. Furthermore, we also develop an effective method to to detect regions undergone contrast enhancement that are different from the rest of the image, and use this method to detect composite images.

A preliminary version of this work was published in [12]. The current work extends our previous method in several key aspects. First, this work uses a more stable property of contrast enhancement on pixel histogram based on the number of empty bins, which leads to a new type of regularizer in the optimization objective. Furthermore, we include a new optimization method based on the Wasserstein distance, which is theoretically better justified. We also augmented the model to handle additive noises. We provide a new and more effective solution based on the projected gradient descent and improve the local estimation algorithm using a new energy minimization formulation. Last but not least, we include all formal proofs to the results in an appendix. This work in the currently submission form has not been previously published, nor is it under consideration for publication elsewhere.

The rest of this paper is organized as follows. In Section 2, we review relevant previous works. In Section 3, we elaborate on the relation of contrast enhancement and pixel histogram, and describe our algorithm estimating parametric and nonparametric contrast enhancement transforms. In section 4, we present the experimental evaluations to the algorithm. Section 5 focuses on a local contrast enhancement estimation algorithm based on an energy minimization framework. Section 6 concludes the article with discussion and future works.

## 2 BACKGROUND AND RELATED WORKS

### 2.1 Contrast Enhancement

We focus on gray-scale images of $b$ bit-pixels. A (normalized) pixel histogram represents the fractions of pixels taking an individual value out of all $2^b$ different values in the image, and is interpreted as the probability distribution of a random variable $X$ over $\{0, \cdots, n = 2^b - 1\}$.

A contrast enhancement is a point-wise monotonic transform between pixel values $i, j \in \{0, \cdots, n\}$, defined as $i = \phi(j) := [m(i)]$, where $m(\cdot): [0, n] \mapsto [0, n]$ is a continuous non-decreasing function, and $[\cdot]$ is the rounding operation that maps a real number to its nearest integer.

There are two categories of contrast enhancement transforms. A *parametric* contrast enhancement can be determined with a set of parameters. An example of parametric contrast enhancement is *gamma correction*,

$$j = \phi_\gamma(i) := \left[ n \left( \frac{i}{n} \right)^\gamma \right],$$  (1)

where $\gamma \geq 0$ is the single parameter. Another example of parametric contrast enhancement is *sigmoid stretching*,

$$j = \phi_{\alpha,\mu}(i) := \left[ n \left( \frac{S\left(\frac{i-n\mu}{n\alpha}\right) - S\left(\frac{-n\mu}{n\alpha}\right)}{S\left(\frac{n(1-\mu)}{n\alpha}\right) - S\left(\frac{-n\mu}{n\alpha}\right)} \right) \right],$$  (2)

where $\alpha > 0$ and $\mu \in [0, 1]$ are two parameters, and $S(x) = \frac{1}{1+\exp(-x)}$ is the sigmoid function. A *nonparametric* contrast enhancement affords no simple parametric form and has to be specified for all $i \in \{0, \cdots, n\}$. An example of nonparametric contrast enhancement is *histogram equalization*, which transforms the pixel histogram of an image to match a uniform distribution over $\{0, \cdots, n\}$.

## 2.2 Previous Works

Except for the case of identity, a contrast enhancement will map multiple input values to a single output value (correspondingly, there will be values to which no input pixel value maps), a result from the pigeonhole principle [13]. This leaves the characterizing peaks and gaps in the pixel histogram after a contrast enhancement is applied, which have inspired several works to develop statistical features to detect the existence of contrast enhancement in an image. The works in[4, 5] describe an iterative algorithm to jointly estimate a gamma correction, based on a probabilistic model of pixel histogram and an exhaustive matching procedure to determine which histogram entries are most likely to correspond to artifacts caused by gamma correction. The statistical procedure of[4] is further refined in [7] to determine if an image has undergone gamma correction or histogram equalization. However, all these methods aim only to detect the existence of certain contrast enhancement in an image, but not to recover the actual form of the contrast enhancement function.

There exist several methods that can also recover the functional form of contrast enhancement. The method in[3] recovers gamma correction from an image using its bi-spectra statistics. The method of[6] uses the the features developed in[5] to recover the actual gamma value by applying different gamma values to a uniform histogram and identifying the optimal value that best matches the observed pixel histogram features. This work is further extended in[8] to recover contrast enhancement of a JPEG compressed image. However, most of the previous methods require knowledge of the type of contrast enhancement a priori. Another common problem of these previous methods is that they are not robust with regards to additive noise that is added to intentionally hide the trace of contrast enhancement[10, 11].

Methods for local contrast enhancement detection have also been studied in several previous works[4, 7, 8]. However, most of these works are limited in that they can only recover at the level of image blocks of large sizes. The high computational cost is one reason that existing methods cannot be used for *pixel-level* localization of different contrast enhancement.

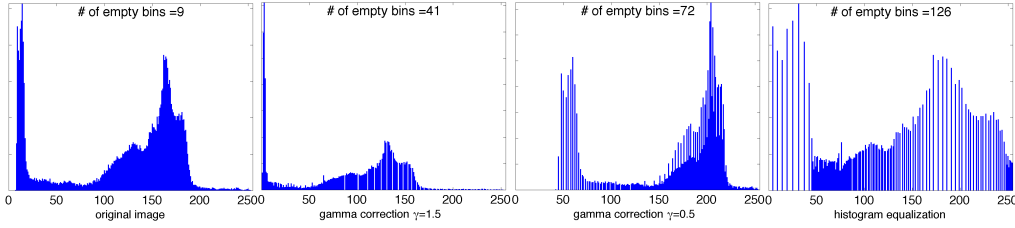

Fig. 1. *Effects of contrast enhancement on the pixel histogram. The four panels correspond to the pixel histograms of (i) the original image, (ii) the image after a gamma correction with $\gamma = 1.5$, (iii) the image after a gamma correction with $\gamma = 0.5$, and (iv) the image after histogram equalization, each panel is marked with the number of empty bins in the corresponding pixel histogram.*

## 3 METHOD

In this section, we describe our method for contrast enhancement estimation from an image. We start with a property of the pixel histogram of contrast enhancement transformed image. This property is then utilized to formulate the contrast enhancement estimation problem as an optimization problem, and we further provide details of the solution algorithm.

### 3.1 Relation of Contrast Enhancement and Pixel Histogram

Consider two images $I$ and $\tilde{I}$, with $\tilde{I}$ obtained from $I$ with a contrast enhancement $\phi$. We denote two random variables $X, Y \in \{0, \cdots, n\}$ as the pixels of $I$ and $\tilde{I}$, hence $Y = \phi(X)$. Using the probability interpretation of pixel histogram, the contrast enhancement between $X$ and $Y$ induces the conditional probability distribution $\Pr(Y = j | X = i) = \mathbf{1}\,(j = \phi(i))$, where $\mathbf{1}(c)$ is the indicator function whose output is 1 if $c$ is true and zero otherwise. As such, we have

$$\Pr(Y = j) = \sum_i \mathbf{1}\,(j = \phi(i)) \Pr(X = i). \tag{3}$$

We can model a pixel histogram as a vector on the $n$-dimensional probability simplex, *i.e.*, $\mathbf{h} \in \Delta^{n+1} := \{\mathbf{h} | \mathbf{h} \succeq 0, \mathbf{1}^T \mathbf{h} = 1\}$ with $h_{i+1} = \Pr(X = i)$ for $i = 0, \cdots, n$, and a contrast enhancement $\phi$ as an $(n + 1) \times (n + 1)$ matrix $T_\phi : (T_\phi)_{i+1, j+1} = \mathbf{1}\,(j = \phi(i))$. Note that $T$ has nonnegative entries, and each column sum to one, *i.e.*, $T^\top \mathbf{1} = \mathbf{1}$. With vector $\mathbf{h}$ and matrix $T_\phi$, we can rewrite Eq. (3) as

$$\tilde{\mathbf{h}} = T_\phi \mathbf{h}, \tag{4}$$

*i.e.*, the pixel histograms of $I$ and $\tilde{I}$ are related by a *linear* transform though $\phi$ is a nonlinear function of $X$.

The contrast enhancement redistributes pixel values in an image. Previous studies have used the characteristic "gaps and peaks" [5] or increased non-smoothness [9] to recover the corresponding contrast enhancement. In this work, we discover a more reliable property of the pixel histogram of contrast enhancement transformed images that leads to better estimation performance, *i.e.*, the number of empty bins in the pixel histogram does not decrease when a contrast enhancement transform is applied to an image. This is formally described in the following result.

THEOREM 1. *Define*

$$\Omega(\mathbf{h}) = \sum_{i=1}^n \mathbf{1}\,(h_i = 0)\,, \tag{5}$$

i.e., $\Omega(\mathbf{h})$ *counts the number of empty bins in* $\mathbf{h}$. *We have* $\Omega(\mathbf{h}) \leq \Omega(T_\phi \mathbf{h})$.

PROOF. We consider a non-empty bin in the pixel histogram of an image. As the contrast enhancement transform transports each bin as a whole (*i.e.*, no splitting of bins), there are only two situation can occur: either this bin becomes another individual non-empty bin in the contrast enhancement transformed image, or it is mapped to a location other bins are also mapped to. In either case, the number of non-empty bins will not increase, and correspondingly the number of empty bins in the pixel histogram after contrast enhancement is applied does not decrease. On the other hand, the strict inequality does not hold, as if we have a monotone image with a single distinct pixel value, any contrast enhancement will only create another monotone image, in which case the number of empty bins stays the same.  □

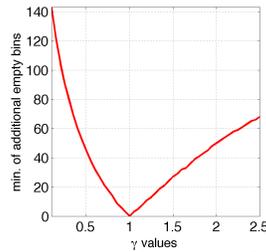

Fig. 2. *The minima of changes in the number of empty bins over* 500 *natural images after gamma corrections with $\gamma$ value in the range of $\{0.1 : 0.01 : 2.5\}$ are applied. Note the prominent trend of increasing number of empty bins after contrast enhancement is applied to these images. See more details in the texts.*

Albeit this is a simple observation, it holds regardless of the image or the contrast enhancement transform. Fig.1 demonstrate this effect using an 8-bit grayscale image as an example. The four panels of Fig.1 correspond to the pixel histograms of (i) the original image, (ii) the image after a gamma correction with $\gamma = 1.5$, (iii) the image after a gamma correction with $\gamma = 0.5$, and (iv) the image after histogram equalization, each with the number of empty bins annotated. Note the significant increment of the number of empty bins in the contrast enhancement transformed images (*e.g.*, the number of empty bins increases from 9 to 126 in the case of histogram equalization).

Fig.2 corresponds to a quantitative evaluation. Specifically, we choose 200 natural images from RAISE dataset[14][1], and apply gamma corrections with $\gamma$ value in the range of $\{0.1 : 0.01 : 2.5\}$ to each image. We then compute the *difference* between the number of empty bins of the gamma corrected image with that of the original image (therefore it is always zero for $\gamma = 1$ which corresponds to the original image). We then show the *minima* of these differences over the 500 images in Fig.2. Note that these minima are positive, indicating a prominent trend of increasing number of empty bins after contrast enhancement is applied.

### 3.2 Effect of Additive Noise

The change in the number of empty bins of pixel histogram caused by contrast enhancement may be obscured by adding noise to the contrast enhancement transformed image (see Fig.3), a fact employed in recent anti-forensic techniques aiming to hiding the trace of contrast enhancement[10, 11]. However, this artifact introduced by additive noise can be precisely modeled using the same mathematical framework. Consider two random variables $X, Y \in \{0, \cdots, n\}$ corresponding to pixel values from two images $I$ and $\tilde{I}$, with $Y = \phi(X) + Z$, where $Z$ is a real-valued white noise with probability density function $p(z)$ and independent

---

[1]The original images are in the 12-bit or 14-bit uncompressed or lossless compressed NEF or TIFF format. We downloaded the full RAISE dataset but use a random subset of 200 images for testing our algorithm. We use the green channel of the RGB color image as in [8]. The pixel histograms are vectors of $2^{12} = 4,096$ and $2^{14} = 16,384$ dimensions, respectively.

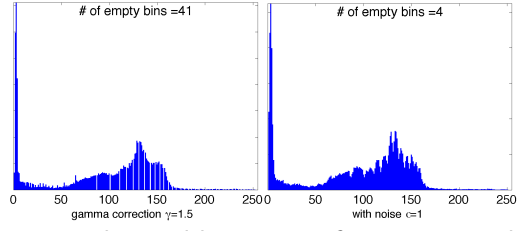

Fig. 3. *Effects of additive noise on the pixel histogram after contrast enhancement is applied. (*Left*): pixel histogram of an image after gamma correction is applied, (*Right*) The same pixel histogram after white Gaussian noise of standard deviation* 1.0 *is added.*

of $X$. Using relations of the probabilities, we have

$$
\begin{aligned}
\Pr(Y = j) &= \sum_i \Pr(Y = j | X = i) \Pr(X = i) \\
&= \sum_i \Pr\left(Z \in [j - \phi(x) - \tfrac{1}{2}, j - \phi(x) + \tfrac{1}{2}]\right) h_{i+1} \\
&= \sum_{i,k} \Pr\left(Z \in [j - k - \tfrac{1}{2}, j - k + \tfrac{1}{2})\right) \mathbf{1}(\phi(i) = k) h_{i+1} \\
&= \sum_{i,k} \Pr\left(Z \in [j - k - \tfrac{1}{2}, j - k + \tfrac{1}{2})\right) (T_\phi)_{i+1,k+1} h_{i+1} \\
&= \sum_{i,k} \left(\int_{j-k-\frac{1}{2}}^{j-k+\frac{1}{2}} p(z) dz\right) (T_\phi)_{i+1,k+1} h_{i+1}
\end{aligned}
$$

We introduce a new matrix

$$
R_{j+1,k+1} = \int_{j-k-\frac{1}{2}}^{j-k+\frac{1}{2}} p(z) dz,
$$

which has a Toeplitz structure and can be obtained in closed form for certain type of noise. For instance, if $Z$ is zero-mean Gaussian noise with standard deviation $\sigma$, elements of matrix $R$ are explicitly given as:

$$
R_{j+1,k+1} = \mathrm{erf}\left(\frac{j - k + \frac{1}{2}}{\sigma}\right) - \mathrm{erf}\left(\frac{j - k - \frac{1}{2}}{\sigma}\right),
$$

$\mathrm{erf}(z) = \int_{-\infty}^{u} \frac{1}{\sqrt{2\pi}} e^{-\frac{\tau^2}{2}} d\tau$ is the cumulative density function of the standardized Gaussian random variable. Subsequently, we assume $R$ is known or can be obtained from images using methods of blind noise estimation, *e.g.*,[15, 16]. Considering the noise effect, the relation between the pixel histograms of the original image and the image after contrast enhancement is applied and white noise added can be expressed as

$$
\tilde{\mathbf{h}} = RT_\phi \mathbf{h}, \tag{6}
$$

which will be used subsequently to recover the contrast enhancement.

## 3.3 Estimating Contrast Enhancement and Pixel Histogram

The problem we are to solve is to recover the unknown contrast enhancement $\phi$ and the pixel histogram of the original image $\mathbf{h}$ simultaneously, using only the pixel histogram of the observed image $\tilde{\mathbf{h}}$.

To this end, we need to measure distances between two histograms. In this work, we use the the Wasserstein distance[17] (also known as Mallows distance or earth mover's distance[18] and subsequently referred to as $W_1$ distance) of two pixel histograms, $\mathbf{h}, \tilde{\mathbf{h}} \in \Delta^{n+1}$, defined as

$$
W_1(\tilde{\mathbf{h}}, \mathbf{h}) = \sum_i \left| \mathcal{C}_{Y \sim \tilde{\mathbf{h}}}(i) - \mathcal{C}_{X \sim \mathbf{h}}(i) \right| = \left\| F\tilde{\mathbf{h}} - F\mathbf{h} \right\|_1, \tag{7}
$$

where $\mathcal{C}_{X \sim \mathbf{h}}(i) = \Pr(X \leq i) = \sum_{j=1}^{i+1} h_j$ is the cumulative distribution of $\mathbf{h}$. Note that $\mathcal{C}_{X \sim \mathbf{h}}$ can be computed as $F\mathbf{h}$, where $F$ is the lower triangular matrix with all one elements.

Using the $W_1$ distance between two pixel histograms and the observation that contrast enhancement leads to nonsmooth pixel histogram, we can formulate contrast enhancement estimation as the following optimization problem

$$\begin{aligned}
\min_{\mathbf{h}, \phi} \quad & W_1(\tilde{\mathbf{h}}, RT_\phi \mathbf{h}) + \lambda \Omega(\mathbf{h}), \\
\text{s.t.} \quad & \mathbf{h} \in \Delta^{n+1}, \phi(i) \le \phi(i+1), i = 0, \cdots, n.
\end{aligned} \tag{8}$$

The first term in the objective function corresponds to relations of pixel histogram of image with and without contrast enhancement in Eq.(6). The second term reflects the observation in Section 3.1 that pixel histogram of the original image tend to have fewer number of empty bins. Parameter $\lambda$ controls the contribution of the two terms in the objective function. The constraint enforces $\mathbf{h}$ to be a legitimate pixel histogram and $\phi$ being a monotonic mapping.

We solve (8) using a block coordinate descent scheme[19] by alternating minimizing the objective with regards to $\mathbf{h}$ and $\phi$ with the other fixed. The estimation of $\mathbf{h}$ with fixed $\phi$ reduces to a convex optimization problem [20], and we describe its solution in Section 3.3.1. The estimation of $\phi$ with fixed $\mathbf{h}$ is then given in Section 3.3.2 for the parametric case and 3.3.3 for the nonparametric case.

*3.3.1 Recovering Pixel Histogram with Fixed Contrast Enhancement.* Using the equivalent definition of Wasserstein distance given in Eq.(7), the problem of finding optimal $\mathbf{h}$ with fixed $\phi$ is obtained from (8) as

$$\begin{aligned}
\min_{\mathbf{h}} \quad & \left\| F\tilde{\mathbf{h}} - FRT_\phi \mathbf{h} \right\|_1 + \lambda \Omega(\mathbf{h}) \\
\text{s.t.} \quad & \mathbf{1}^T \mathbf{h} = 1, \mathbf{h} \succeq 0.
\end{aligned} \tag{9}$$

Eq.(9) is difficult to solve because (i) the $W_1$ distance uses the non-differentiable $\ell_1$ norm and (ii) $\Omega(\mathbf{h})$ is not a continuous function. To proceed, in this work, we replace the non-differentiable $\| \cdot \|_1$ using a generalization of the result in [21] (Theorem 2, proof in Appendix 6).

THEOREM 2. *For* $\mathbf{x} \in \mathcal{R}^n$, *we have*

$$\|\mathbf{x}\|_1 = \min_{\mathbf{z} \succeq 0} \frac{1}{2} \left( \mathbf{x}^\top \mathcal{D}(\mathbf{z})^{-1} \mathbf{x} + \mathbf{1}^\top \mathbf{z} \right), \tag{10}$$

*and*

$$|\mathbf{x}| = \operatorname*{argmin}_{\mathbf{z} \succeq 0} \frac{1}{2} \left( \mathbf{x}^\top \mathcal{D}(\mathbf{z})^{-1} \mathbf{x} + \mathbf{1}^\top \mathbf{z} \right). \tag{11}$$

$\mathcal{D}(\mathbf{z})$ *denotes a diagonal matrix formed from vector* $\mathbf{z}$ *as its main diagonal, and* $|\mathbf{x}|$ *as the vector formed from the absolute values of the components of* $\mathbf{x}$.

Furthermore, note that for $\rho > 0$, we have $e^{-\rho c} \ge \mathbf{1}(c = 0)$ with equality holding iff $c = 0$, and for $c > 0$, $e^{-\rho c} \to 0$ with $\rho \to \infty$. This means that we can use $e^{-\rho \mathbf{h}}$, where the exponential is applied to each element of vector $\mathbf{h}$, as a continuous and convex surrogate to the non-continuous function in $\Omega(\mathbf{h})$.

Using these results, we can develop an efficient numerical algorithm. First, we introduce an auxiliary variable $\mathbf{u} \succeq 0$ to replace the $\ell_1$ norms in (9), and use the scaled exponential function to reformulate the problem as

$$\min_{\mathbf{h}, \mathbf{u}} L(\mathbf{h}, \mathbf{u}), \text{ s.t. } \mathbf{1}^T \mathbf{h} = 1, \mathbf{h} \succeq 0, \mathbf{u} \succeq 0, \tag{12}$$

where $L(\mathbf{h}, \mathbf{u})$ equals to

$$\frac{1}{2} \left( \tilde{\mathbf{h}} - RT_\phi \mathbf{h} \right)^\top F^\top \mathcal{D}(\mathbf{u})^{-1} F \left( \tilde{\mathbf{h}} - RT_\phi \mathbf{h} \right) + \frac{1}{2} \mathbf{1}^\top \mathbf{u} + \lambda e^{-\rho \mathbf{h}}.$$

This is a convex constrained optimization problem for $(\mathbf{h}, \mathbf{u})$. While in principle we can use off-the-shelf convex programming packages such as `CVX` [22] to solve this problem, the

---

**ALGORITHM 1:** Optimization of (9)

---

initialize $\mathbf{h}_0$ and $\mathbf{u}_0$, $t \leftarrow 0$
**while** not converge **do**
    $\mathbf{h}_{t,0} \leftarrow \mathbf{h}_t$, $\tau \leftarrow 0$
    **while** not converge **do**
        $M \leftarrow (FRT_\phi)^\top \mathcal{D}(\mathbf{u}_t)^{-1} (FRT_\phi)$
        $\mathbf{b} \leftarrow (FRT_\phi)^\top \mathcal{D}(\mathbf{u}_t)^{-1} F\tilde{\mathbf{h}}$
        $\triangle\mathbf{h}_\tau \leftarrow M\mathbf{h}_{t,\tau} - \mathbf{b} - \lambda\rho e^{-\rho\mathbf{h}_{t,\tau}}\mathbf{h}_{t,\tau}$
        $\mathbf{h}_{t,\tau+1} \leftarrow \mathcal{P}_{\Delta^{n+1}}\left(\mathbf{h}_{t,\tau} - \frac{\eta_0}{\tau+1}\triangle\mathbf{h}_\tau\right)$
        $\tau \leftarrow \tau + 1$
    **end while**
    $\mathbf{h}_{t+1} \leftarrow \mathbf{h}_{t,\tau}$, $\mathbf{u}_{t+1} \leftarrow \left| F\left(\tilde{\mathbf{h}} - RT_\phi \mathbf{h}_{t+1}\right) \right|$
    $t \leftarrow t + 1$
**end while**

---

potential high dimensionality of $\mathbf{h}$ (for instance for 14-bit image $\mathbf{h}$ is of $16,384$ dimensions) requires a more efficient algorithm designed for our problem.

To this end, we solve (12) with a block coordinate descent sub-procedure by iterating $\mathbf{h}_{t+1} \leftarrow \text{argmax}_{\mathbf{h}} L(\mathbf{h}, \mathbf{u}_t)$ and $\mathbf{u}_{t+1} \leftarrow \text{argmax}_{\mathbf{u}} L(\mathbf{h}_{t+1}, \mathbf{u})$ until convergence. The overall algorithm minimizing (9) is summarized in pseudocode in Algorithm 1.

**Optimizing $\mathbf{h}$**: fixing $\mathbf{u}_t$ and dropping irrelevant terms, minimizing $\mathbf{h}$ reduces to the following constrained nonlinear convex optimization problem

$$\begin{aligned} \min_{\mathbf{h}} \quad & \tfrac{1}{2}\mathbf{h}^\top M\mathbf{h} - \mathbf{b}^\top \mathbf{h} + \lambda e^{-\rho\mathbf{h}} \\ \text{s.t.} \quad & \mathbf{1}^T\mathbf{h} = 1, \mathbf{h} \succeq 0, \end{aligned} \tag{13}$$

where

$$\begin{aligned} M \;&= (FRT_\phi)^\top \mathcal{D}(\mathbf{u}_t)^{-1} (FRT_\phi) \\ \mathbf{b} \;&= (FRT_\phi)^\top \mathcal{D}(\mathbf{u}_t)^{-1} F\tilde{\mathbf{h}}. \end{aligned}$$

Eq.(13) can be efficiently solved with a projected gradient descent method [20]. Specifically, starting with $\mathbf{h}_{t,0} = \mathbf{h}_t$, our algorithm iterates between two steps:

$$\begin{aligned} \triangle\mathbf{h}_\tau \;&\leftarrow M\mathbf{h}_{t,\tau} - \mathbf{b} - \lambda\rho e^{-\rho\mathbf{h}_{t,\tau}}\mathbf{h}_{t,\tau}, \\ \mathbf{h}_{t,\tau+1} \;&\leftarrow \mathcal{P}_{\Delta^{n+1}}\left(\mathbf{h}_{t,\tau} - \frac{\eta_0}{\tau+1}\triangle\mathbf{h}_\tau\right). \end{aligned} \tag{14}$$

The first step computes the gradient of the objective function. The second equation performs a gradient descent update with step size $\frac{\eta_0}{\tau+1}$ followed by a projection onto the $n$-dimensional probability simplex $\mathcal{P}_{\Delta^{n+1}}$, which does not have closed-form but affords a very efficient algorithm (see Appendix 6 for a detailed description for completeness). The damping step size $\frac{\eta_0}{\tau+1}$ guarantees the convergence of the projected gradient descent algorithm [20], and we choose $\eta_0 = 1.2$ in all our subsequent experiments. This projected gradient descent algorithm usually converges within 5-10 steps. We take $\mathbf{h}_{t+1} = \mathbf{h}_{t,\tau}$ at the convergence.

**Optimizing $\mathbf{u}$**: when fixing $\mathbf{h}_{t+1}$ and dropping irrelevant terms, minimizing $\mathbf{u}$ becomes

$$\min_{\mathbf{u}} \mathbf{c}^\top \mathcal{D}(\mathbf{u})^{-1}\mathbf{c} + \mathbf{1}^\top\mathbf{u} \text{ s.t. } \mathbf{u} \succeq 0,$$

where $\mathbf{c} = F\left(\tilde{\mathbf{h}} - RT_\phi\mathbf{h}_{t+1}\right)$. Using Theorem 2, we obtain the optimal solution $\mathbf{u}_{t+1} = |\mathbf{c}|$.

---

**ALGORITHM 2:** Estimation of Parametric Contrast Enhancement

---

    **for** $\theta \in \{\theta_1, \cdots, \theta_m\}$ **do**
        compute $L^\star(\theta)$ using Algorithm 1;
    **end for**
    return $\theta^\star = \text{argmin}_{\theta \in \{\theta_1, \cdots, \theta_m\}} L^\star(\theta)$

---

*3.3.2 Estimating Parametric Contrast Enhancement.* For a parametric contrast enhancement transform that can be determined by small set of parameters $\theta$, *e.g.*, gamma correction (1) (where $\theta = \gamma$) and sigmoid stretching (2) (where $\theta = (\alpha, \mu)$), even though the parameters are continuous, the discrete nature of contrast enhancement as transforms between integers means that there are only finite number of distinguishable parameter values. We illustrate this point in the case of gamma correction. Consider the 2D lattice $\{0, \cdots, n\} \times \{0, \cdots, n\}$, the graph of gamma correction corresponds to a path over grid points $(i, j)$ starting from $(0, 0)$ and ending at $(n, n)$. This path is monotonic, *i.e.*, it never dips down. Furthermore, for $\gamma < 1$, the path stays on or above the diagonal, while for $\gamma > 1$, the path stays on or below the diagonal. These properties give rise to only a finite set of distinguishable $\gamma$ values, (*i.e.*, values leading to different gamma correction transforms) as the following result shows (proved in Appendix 6).

THEOREM 3. *All* $\gamma \in [\underline{\gamma}_{ij}, \bar{\gamma}_{ij})$ *for* $i, j \in \{1, \cdots, n-1\}$ *where*

$$\underline{\gamma}_{ij} = \frac{\log n - \log(j + \frac{1}{2})}{\log n - \log i}, \bar{\gamma}_{ij} = \frac{\log n - \log(j - \frac{1}{2})}{\log n - \log i}.$$

*leads to the same gamma correction curve. As such, the total number of distinguishable* $\gamma$ *value is bounded by* $(n-1)^2$.

In practice, distinguishable parameter values are also limited by the numerical precision in which they can be input in photo editing software, usually in the range of $10^{-2}$ or $10^{-3}$.

On the other hand, optimal contrast enhancement parameters lead to a minimum of Eq.(12) across different parameter values. As the transformed pixel histogram will be exactly the same as the observed histogram, thus the first term will reach minimum (zero), while the original histogram should have the maximum degree of smoothness reflected by a minimal total variation. We denote the minimum of (12) corresponding to contrast enhancement parameter $\theta$ as $L^\star(\theta)$. Fig.4 shows the graph of $L^\star(\theta)$ for the case of gamma correction. The curves correspond to two different parameter values: $\gamma = 0.4$ and $1.4$, and range of searched $\gamma$ values is $[0.1, 1.8]$ with a stride of $0.05$. In both cases, the true $\gamma$ values lead to global minimums of $L^\star(\theta)$[2].

These two characteristics of parametric contrast enhancement, *i.e.*, there are finite number of distinguishable parameter values and the optimal value usually corresponds to a global minimum of (12) suggests that the optimal parameter can be recovered by a grid search in the set of plausible parameters. Specifically, given a search range of parameter values $\Phi = \{\theta_1, \cdots, \theta_m\}$, we seek $\theta^\star = \text{argmin}_{\theta \in \Phi} L^\star(\theta)$ as the optimal contrast enhancement parameter. This is the algorithm we use for estimating parametric contrast enhancement. A pseudo code is given in Algorithm 2.

---

[2]Similar observations have also been made on other types of parametric contrast enhancement transforms, such as sigmoid stretching and cubic spline curves.

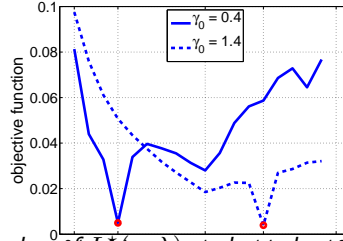

Fig. 4. *The relation between the value of $L^\theta(\gamma, \lambda)$ and $\gamma$ values in the case of contrast mapping in the form of simple gamma correction. The true $\gamma$ value in the two cases are $0.4$ and $1.4$, which correspond to the global minima of the two curves.*

---

**ALGORITHM 3:** Optimization of (17)

---

initialize $\hat{\mathbf{h}}_0$, $\hat{\mathbf{u}}_0$ and $\hat{\mathbf{v}}_0$, $t \leftarrow 0$
**while** not converge **do**
    $\hat{\mathbf{h}}_{t,0} \leftarrow \hat{\mathbf{h}}_t$, $\tau \leftarrow 0$
    **while** not converge **do**
        $\hat{M} \leftarrow (RF)^\top \mathcal{D}(\hat{\mathbf{u}}_t)^{-1} RF + \xi F^\top \mathcal{D}(\hat{\mathbf{v}}_t)^{-1} F$
        $\hat{\mathbf{b}} \leftarrow (RF)^\top \mathcal{D}(\hat{\mathbf{u}}_t)^{-1} F\tilde{\mathbf{h}} + \xi F^\top \mathcal{D}(\hat{\mathbf{v}}_t)^{-1} F T_\phi \mathbf{h}$
        $\triangle \hat{\mathbf{h}}_\tau \leftarrow \hat{M} \hat{\mathbf{h}}_{t,\tau} - \hat{\mathbf{b}}$
        $\hat{\mathbf{h}}_{t,\tau+1} \leftarrow \mathcal{P}_{\Delta^{n+1}} \left( \hat{\mathbf{h}}_{t,\tau} - \frac{\eta_0}{\tau+1} \triangle \hat{\mathbf{h}}_\tau \right)$, $\tau \leftarrow \tau + 1$
    **end while**
    $\hat{\mathbf{h}}_{t+1} \leftarrow \hat{\mathbf{h}}_{t,\tau}$, $\hat{\mathbf{u}}_{t+1} \leftarrow \left| F\left( \tilde{\mathbf{h}} - R\hat{\mathbf{h}}_{t+1} \right) \right|$;
    $\hat{\mathbf{v}}_{t+1} = \left| F\left( \hat{\mathbf{h}}_{t+1} - T_\phi \mathbf{h} \right) \right|$; $t \leftarrow t + 1$;
**end while**

---

### 3.3.3 Estimating Nonparametric Contrast Enhancement.
In the case where the contrast enhancement transform does not afford a parametric form, with the pixel histogram of the original image obtained using the algorithm given §3.3.2, we estimate $\phi$ directly. Dropping irrelevant terms from the overall optimization problem (8), this reduces to the following problem

$$\min_\phi \quad W_1(\hat{\mathbf{h}}, RT_\phi \mathbf{h}) \tag{15}$$
$$\text{s.t.} \quad \phi(i) \leq \phi(i+1), i = 0, \cdots, n.$$

To solve this problem, we first decouple pixel histogram before and after noise is added. To this end, we introduce an auxiliary variable $\hat{\mathbf{h}}$ and a parameter $\xi$ to enforce the constraint $\hat{\mathbf{h}} = T_\phi \mathbf{h}$ with $W_1$ distance, and rewrite the optimization problem as

$$\min_{\hat{\mathbf{h}}, \phi} \quad W_1(\tilde{\mathbf{h}}, R\hat{\mathbf{h}}) + \xi W_1(\hat{\mathbf{h}}, T_\phi \mathbf{h}) \tag{16}$$
$$\text{s.t.} \quad \hat{\mathbf{h}} \in \Delta^{n+1}, \phi(i) \leq \phi(i+1), i = 0, \cdots, n.$$

Using the block coordinate descent scheme, we solve (18) by alternating minimization of $\hat{\mathbf{h}}$ and $\phi$ with the other fixed until the guaranteed convergence is reached.

**Optimizing $\hat{\mathbf{h}}$** This problem becomes

$$\min_{\hat{\mathbf{h}}, \phi} \quad \left\| F\hat{\mathbf{h}} - FR\hat{\mathbf{h}} \right\|_1 + \xi \left\| F\hat{\mathbf{h}} - FT_\phi \mathbf{h} \right\|_1 \tag{17}$$
$$\text{s.t.} \quad \mathbf{1}^T \hat{\mathbf{h}} = 1, \hat{\mathbf{h}} \succeq 0.$$

Following a similar procedure as for the solution of Eq.(9), we optimize (17) with an iterative algorithm (Algorithm 3) that uses the $\ell_1$ relaxation in Theorem 2 and projected gradient descent. We defer a detailed description to the Appendix 6.

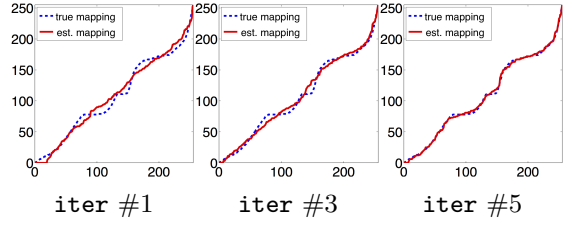

Fig. 5. *Convergence of the estimated nonparametric free-form contrast enhancement transform with Algorithm 4.*

---

**ALGORITHM 4:** Estimation of Nonparametric Contrast Enhancement

---

**while** not converge **do**
    update $\tilde{\mathbf{h}}$ (9) using Algorithm 1;
    update $\hat{\mathbf{h}}$ (17) using Algorithm 3;
    update $\phi$ (18) with (20);
**end while**

---

**Optimizing** $\phi$. We use the equivalent definition of $W_1$ distance based on the cumulative distributions (7), and the problem of solving $\phi$ reduces to

$$\begin{aligned}
\min_\phi \quad & \sum_{i=0}^n \left| \mathcal{C}_{\hat{X}\sim\hat{\mathbf{h}}}(i) - \mathcal{C}_{\phi(X)\sim T_\phi\mathbf{h}}(i) \right| \\
\text{s.t.} \quad & \phi(i) \le \phi(i+1), i = 0, \cdots, n.
\end{aligned} \tag{18}$$

This is essentially the search for a monotonic transform between two random variables $X$ and $\hat{X}$ over $\{0, \cdots, n\}$ with corresponding probability distributions (histograms) $\mathbf{h}$ and $\hat{\mathbf{h}}$, such that the histogram of $\phi(X)$ is as close as possible to that of $\hat{X}$.

The problem of finding a transform that matches random variable of one distribution to another is known as *histogram matching*, the optimal solution of which can be obtained from the cumulative distributions of the two random variables [17]. Specifically, from the cumulative probability distribution of $\mathbf{h}$, $\mathcal{C}_{X\sim\mathbf{h}} : \{0, \cdots, n\} \mapsto [0, 1]$, we define the corresponding pseudo inverse cumulative distribution function $[0, 1] \mapsto \{0, \cdots, n\}$, as

$$\mathcal{C}_{X\sim\mathbf{h}}^{-1}(j) = i \text{ if } \mathcal{C}_{X\sim\mathbf{h}}(i-1) < j \le \mathcal{C}_{X\sim\mathbf{h}}(i). \tag{19}$$

The histogram matching transform is formed by applying the cumulative distribution function of $\mathbf{h}$ followed by the pseudo inverse cumulative distribution function of $\tilde{\mathbf{h}}$ (19),

$$\phi^\star(i) = \mathcal{C}_{Y\sim\tilde{\mathbf{h}}}^{-1}(\mathcal{C}_{X\sim\mathbf{h}}(i)). \tag{20}$$

It can be shown that this function is monotonic and maps $X \sim \mathbf{h}$ to $\phi^\star(X) \sim \tilde{\mathbf{h}}$ and leads to the objective function in (18) to zero.

We provide the pseudo code of the overall algorithm in Algorithm 4. Fig.5 demonstrate the convergence of one estimated contrast enhancement transform, with the original transform obtained from interpolating manually chosen key points using cubic splines. As it shows, after 5 iterations of the algorithm, the estimated transform is already very close to the true transform.

## 4 EXPERIMENTAL EVALUATION

In this section, we report experimental evaluations of the contrast enhancement estimation method described in the previous section. The images used in our experiments are based on $N = 200$ grayscale images from the RAISE dataset[14][3]. All subsequent running time

---

[3]The original images are in the 12-bit or 14-bit uncompressed or lossless compressed NEF or TIFF format. We downloaded the full RAISE dataset but use a random subset of 200 images for testing our algorithm. We

statistics are based on a machine of 3.2GHz Due Core Intel CPU and 16G RAM and unoptimized MATLAB code.

## 4.1 Parametric Contrast Enhancement Estimation

*4.1.1 Gamma Correction.* We first consider the estimation gamma correction transform. We choose $\gamma$ value from the range of $\{0.1 : 0.05 : 2.5\}$, with each $\gamma$ value applied to all $N$ images to create sets of gamma corrected images. We implement the grid-search based algorithm (Algorithm 2 in §3.3.2) using a probing range of $\{0.1 : 0.01 : 2.5\}$ to recover the $\gamma$ values from these images. We use a stride of 0.01 as it is the minimum numerical precision a user can specify a gamma correction in image editing tools. We choose parameter $\lambda = 0.75$ and $\rho = 1$ as we found the estimation results are not particularly sensitive to the values of these parameters. Unless specified, we set the noise level to $\sigma = 0.01$.

We use the estimation accuracy rate (AR) to quantify the estimation performance. For an error threshold $\epsilon$, $A_\epsilon$ corresponds to the fraction of estimations that are within a relative error of $\epsilon$. Specifically, denoting the true parameter as $\gamma^\star$ and the estimated parameter as $\gamma_i$ for each of the $N$ test images, AR is defined as

$$A_\epsilon = \frac{\sum_{i=1}^{N} \mathbf{1}(|\gamma_i - \gamma^\star| \leq \epsilon)}{N}. \tag{21}$$

For a given $\epsilon$, higher AR $A_\epsilon$ corresponds to better estimation performance. Subsequently, we report $A_0$, $A_{0.01}$ and $A_{0.05}$, corresponding to the cases when the estimation is exact, has a relative error $\leq 0.01$ and has a relative error $\leq 0.05$, respectively.

We apply our estimation algorithm and compare it with two previous works on gamma correction estimation[3, 8][4]. The results for $A_0$, $A_{0.01}$ and $A_{0.05}$ for the full range of probing $\gamma$ values are shown in Fig.6. The bi-spectra based method of [3] demonstrates some stable estimation results for $\gamma$ value near 1.0, yet the performance deteriorates as $\gamma$ deviates from 1.0. This may be due to the fact that estimations of bi-spectral features become less reliable for more extreme $\gamma$ values. The original method of [8] is a classification scheme based on the empty-bin locations as classification features. To apply it to the estimation problem, we build 249 classifiers corresponding to the probing range of $\gamma$ values, and output the $\gamma$ value that corresponds to the largest classification score. Using only the locations of the empty bins may not be sufficient to recover the $\gamma$ value as many neighboring $\gamma$ values share similar empty bin locations. On the other hand, our method achieves significant improvement in performance when comparing with those of the two previous works. We attribute the improved estimation performance of our detection algorithm to the fact that the optimization formulation of the problem better captures characteristics of pixel histogram, with the Wasserstein loss reflects different locations of the empty bin, and the regularizer favoring smaller number of empty bins further reduces uncertainty in determining the $\gamma$ value. The average running time is 23.1 second per test image.

*4.1.2 Robustness under Noise and JPEG.* We further evaluate the robustness of our method in the presence of noise and JPEG compression. We apply gamma correction with $\gamma$ values randomly sampled from the range $[0.1, 2.5]$ to generate 100 test images. Then, white Gaussian noises with zero mean and various levels are added to the gamma corrected image and then

---

use the green channel of the RGB color image as in [8]. The pixel histograms are vectors of $2^{12} = 4,096$ and $2^{14} = 16,384$ dimensions, respectively.

[4]We use our own MATLAB implementation of these methods following the settings provided in the corresponding published papers. We compared only with the results of[8] which improves on the earlier work from the same authors in[6].

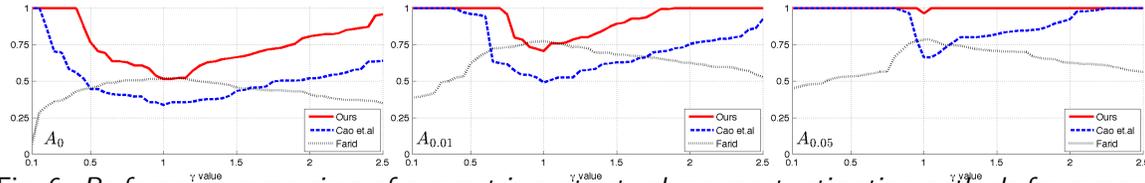

Fig. 6. *Performance comparison of parametric contrast enhancement estimation methods for gamma correction, the three plots corresponds to $A_0$, $A_{0.01}$ and $A_{0.05}$, respectively.*

Table 1. *Robustness under additive Gaussian noise and JPEG compression for the estimation of gamma correction evaluated by $A_{0.05}$.*

|  |  | accuracy | | |
|---|---|---|---|---|
|  |  | ours | [3] | [8] |
| no perturbation | | 98/100 | 64/100 | 82/100 |
| noise | 0.1 | 90/100 | 50/100 | 67/100 |
|  | 0.5 | 81/100 | 31/100 | 58/100 |
|  | 1.0 | 62/100 | 12/100 | 41/100 |
|  | 2.0 | 23/100 | 2/100 | 19/100 |
|  | 5.0 | 11/100 | 0/100 | 8/100 |
| JPEG | 90 | 84/100 | 53/100 | 81/100 |
|  | 80 | 76/100 | 43/100 | 73/100 |
|  | 70 | 69/100 | 34/100 | 64/100 |
|  | 60 | 55/100 | 22/100 | 59/100 |
|  | 50 | 41/100 | 15/100 | 52/100 |

rounded to integral pixel values. We further applied JPEG compression of different quality factors to the same set of gamma corrected images.

Shown in Table 1 are the performance evaluated with $A_{0.01}$, which is the percentage of estimated $\gamma$ that fall in the range of $\pm 0.05$ of the ground truths, as well as comparisons with the methods of[3] and[8]. Accuracies of all methods are affected by the additive noises and JPEG compression. But in the case of noise, the performances of our method show less degradation in comparison with those of the previous works because our method directly incorporate noise perturbations, while the previous works are based on properties that are fragile in the presence of perturbations. On the other hand, in the presence of JPEG compression, our method achieves comparable performance with the method of [8] that is specifically designed to model the artifacts introduced by JPEG to an image after contrast enhancement.

## 4.2 Nonparametric Contrast Enhancement

We further test our methods to recover two different types of nonparametric contrast enhancement transforms – histogram equalization and a nonparametric contrast enhancement created by cubic spline interpolation of manually selecting key points. The latter is analogous to free-form contrast enhancement transform in photo-editing software (*e.g.*, the `Curve` tool in `Photoshop`). We applied histogram equalization and interpolated contrast enhancement transform to create 100 test images of each type.

Because of the nonparametric nature of the contrast enhancement transform, we measure the performance using a slightly different metric based on the relative root mean squared error (RMSE) between $\phi^\star$ and $\phi_k$ to evaluate the performance. Denote the true contrast enhancement transform as $\phi^{\star}$ and the estimated contrast enhancement transform using each test image as $\phi_k$, for an error threshold $\epsilon$, we define the accuracy rate as

$$\hat{A}_{\epsilon} = \frac{1}{N} \sum_{k=1}^{N} \mathbf{1} \left( \frac{\|\phi_k - \phi^{\star}\|_2}{\|\phi^{\star}\|_2} \leq \epsilon \right). \tag{22}$$

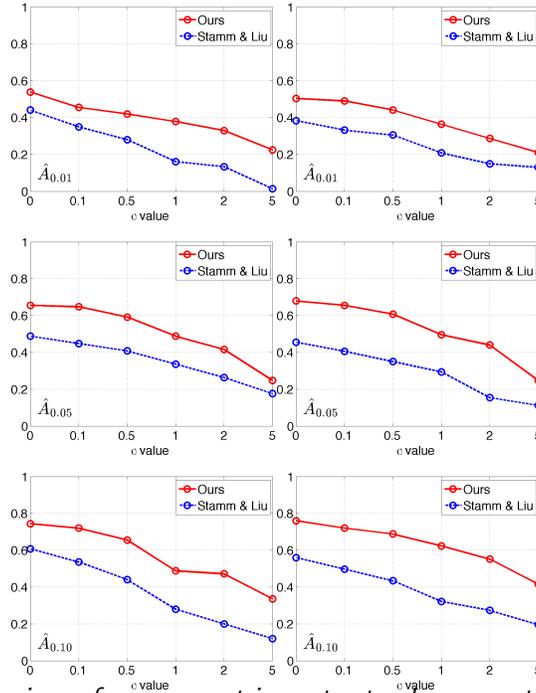

Fig. 7. *Performance comparison of nonparametric contrast enhancement estimation methods for histogram equalization (left) and the freeform contrast enhancement curve, the three rows corresponding to plots of $\hat{A}_{0.01}$, $\hat{A}_{0.05}$ and $\hat{A}_{0.10}$, respectively.*

We implemented our algorithm to recover nonparametric contrast enhancement as described in Algorithm 4 in Section 3.3.3, and set $\lambda = 0.75$ and $\xi = 10$. In practice, we observe that the algorithm usually converges within less than 10 iterations. We compare with the only known previous work for the same task in[4], which is based on an iterative and exhaustive search of pixel histogram that can result in the observed pixel histogram of an image after the contrast enhancement is applied[5].

Fig.7 shows the performance of both algorithms measured by $\hat{A}_{0.01}$, $\hat{A}_{0.05}$ and $\hat{A}_{0.10}$, with perturbations from additive white Gaussian noise of different strengths. As these results show, the estimation performance with our method is consistently better than that of[4]. Furthermore, our method takes about 2 seconds to run on an $800 \times 800$ image and on average it is $5 - 10$ times faster than the method of[4], which relies on an exhaustive search. More importantly, the performances of the method in[4] seem to be strongly affected by noise and compression, this is in direct contrast to our method, which can take such perturbation into consideration to become more robust.

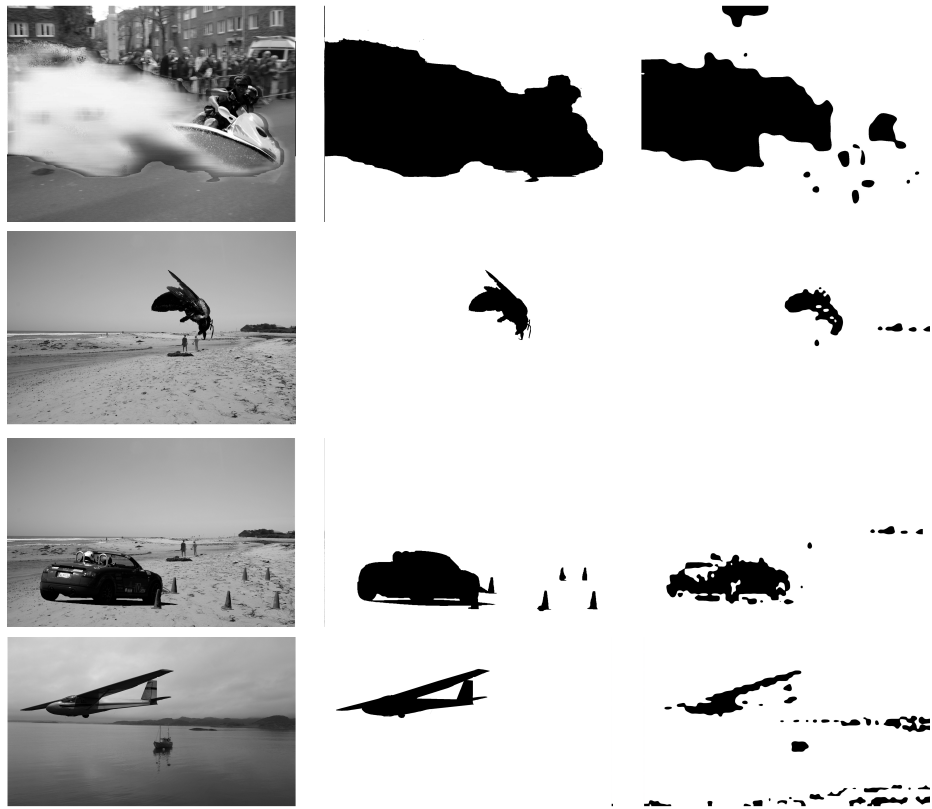

Fig. 8. *Local contrast enhancement estimation for tampering detection.* (**Left**): *Four examples of the manipulated images created from the NIMBLE dataset.* (**Middle**): *Ground truth masks of the spliced regions.* (**Right**): *The detection results with black and white corresponding to regions with label* 0 *and* 1.

## 5   LOCAL CONTRAST ENHANCEMENT ESTIMATION

A composite image can be created by replacing a region in one image with a region from a different image with the aim to modify the content of the original image. If the two images used to create the composite image were captured under different lighting environments, an image forger may need to perform contrast enhancement so that lighting conditions match across the composite image. Image forgeries created in this manner can thus be identified by using a local contrast enhancement estimation method that reveals inconsistent contrast enhancement transforms across different regions.

A straightforward approach would be to apply the global contrast enhancement detection method to non-overlapping rectangular blocks of pixels[4, 7, 8]. However, this simple method has several problems. First, due to the smaller number of pixels in each block, it is difficult to obtain a reliable estimation of the contrast enhancement transform. Second, this simple approach does not take into consideration that adjacent blocks are likely to have undergone the same contrast enhancement. The third problem with these methods is that, to avoid long running time, these methods are only run on non-overlapping blocks and obtain *block level* localization, while practical forensic analysis it is desirable to have *pixel level* localization of regions undergone different contrast enhancement. To improve on these aspects, in this section, we describe a new local contrast enhancement estimation method based on our global estimation method, but embed it in an energy minimization formulation for a more effective and efficient local contrast enhancement estimation at the pixel level.

---

[5]Code of this work is not made public by the authors, so the results are based on our own implementation using `MATLAB` following the descriptions in[4].

## 5.1 Energy Minimization

We segment an image into $m$ overlapping blocks, $\{I_1, \cdots, I_m\}$, and denote $\mathscr{N}(k) \subseteq \{1, \cdots, m\}$ as the indices of blocks that are spatial neighbors of block $I_k$ based on a 4-connected neighborhood system. We use an operator $\mathcal{H}(I)$ to denote the procedure of obtaining pixel histogram from an image region $I$.

We assume that there are two regions in the image undergone two different and unknown contrast enhancement transforms, $\phi_0$ and $\phi_1$, and associate each block with a binary label $y_k \in \{0, 1\}$: $y_k = 0$ indicating that $\phi_0$ is applied to $I_k$ and $y_k = 1$ indicating $I_k$ has contrast enhancement $\phi_1$[6]. Our algorithm obtains estimation of $\phi_0$ and $\phi_i$ and image regions to which they are applied simultaneously, which is formulated as minimizing the following energy function with regards to $\phi_0, \phi_1$ and labels $\{y_k\}_{k=1}^m$:

$$\sum_k \mathcal{E}_{y_k}(I_k) + \beta \sum_k \sum_{k' \in \mathcal{N}(k)} |y_k - y_{k'}|. \qquad (23)$$

The first term in Eq.(23) is the unary energy or data term for the two values of the label, which is defined as

$$\begin{aligned} \mathcal{E}_0(I) &= \min_{\mathbf{h} \in \Delta^{n+1}} \log \left( W_1(\mathcal{H}(I), RT_{\phi_0}\mathbf{h}) + \lambda\Omega(\mathbf{h}) \right) \\ \mathcal{E}_1(I) &= \min_{\mathbf{h} \in \Delta^{n+1}} \log \left( W_1(\mathcal{H}(I), RT_{\phi_1}\mathbf{h}) + \lambda\Omega(\mathbf{h}) \right). \end{aligned} \qquad (24)$$

Note that the unary energy is obtained as the result of optimizing Eq.(9) assuming known contrast enhancement, thus it can be computed with Algorithm 1.

The second term in Eq.(23), $\beta \sum_k \sum_{k' \in \mathcal{N}(k)} |y_k - y_{k'}|$, corresponds to the binary energy that penalizes differences of label assignments to neighboring blocks. It reflects the assumption that the same contrast enhancement is applied to an extended region in the image that subsumes many neighboring blocks. Parameter $\beta$ is used to balance the numerical contribution of the unary and binary energy in the overall energy function (we use $\beta = 0.1$ in the subsequent experiments).

The minimization of Eq.(23) is a mixed optimization problem with discrete labels and continuous functions, and we solve it by an iterative block coordinate descent algorithm that alternates between the optimization of $(\phi_0, \phi_1)$ and $\{y_k\}_{k=1}^m$ with the other set of variable fixed.

**Optimizing** $\{y_k\}_{k=1}^m$. With fixed $\phi_0$ and $\phi_1$, the unary energy in Eq.(24) can be computed with Algorithm 1. The energy function is a sub-modular function of the binary labels $\{y_k\}_{k=1}^m$, which can be minimized using the graph cut algorithm [23].

**Optimizing** $(\phi_0, \phi_1)$. The update for $\phi_0$ and $\phi_1$ with fixed cluster labels proceeds as re-estimating $\phi_0$ and $\phi_1$ using the union of all pixels in blocks with the corresponding label equal to 0 and 1, respectively, as

$$\begin{aligned} \phi_0 &\leftarrow \operatorname{argmin}_\phi \min_{\mathbf{h}} W_1(\mathcal{H}(\cup_{k:y_k=0} I_k), RT_{\phi_0}\mathbf{h}) + \lambda\Omega(\mathbf{h}) \\ \phi_1 &\leftarrow \operatorname{argmin}_\phi \min_{\mathbf{h}} W_1(\mathcal{H}(\cup_{k:y_k=1} I_k), RT_{\phi_0}\mathbf{h}) + \lambda\Omega(\mathbf{h}) \end{aligned} \qquad (25)$$

---

[6]The subsequent algorithm can be readily extended to the case of more than two regions by replacing the binary label with multi-valued labels.

This is implemented as first collecting pixel histogram on these blocks, then apply Algorithm 3 in Section 3.3.3 to recover the contrast enhancement transform[7]. Compared to previous methods that are based on single image blocks, this increases the number of pixels in the estimation of pixel histogram and thus improves the stability of the estimation.

## 5.2 Experimental Evaluation

We perform experimental evaluations of the local contrast enhancement estimation algorithm using a set of 100 composite images with ground truth masks of the spliced regions. These images are a subset of the NIMBLE Challenge dataset provided by NIST for evaluating existing image forensic methods[8]. We applied different contrast enhancement transforms to the spliced regions. To ensure diversity of the applied contrast enhancement, we choose from four different cases, including gamma correction, sigmoid stretching, histogram equalization and monotonic cubic spline curve with hand picked control points. Four examples of the manipulated images are shown in the left column of Fig.8, with the ground truth masks of spliced region shown in the middle column.

We implement the local contrast enhancement estimation algorithm described in the previous section. We try to simulate a situation where the specific form of contrast enhancement is unknown, so we use Algorithm 3 in Section 3.3.3 and ignore the fact that some of the contrast enhancements have parametric form. The size and stride of these blocks determine the reliability of the estimated contrast enhancementtransforms and the accuracy in locating the spliced regions in a composite image. Empirically, we found block size of $50 \times 50$ pixels with overlapping strides of 2 pixels provide a good tradeoff of running efficiency and estimation accuracy, so they are used throughout our subsequent experiments.

The detection results on the four examples given in the leftmost column of Fig.8 are shown in the corresponding panels on the right, with black and white corresponding to regions with label 0 and 1. To quantitatively evaluating the results, we use the detection ratio ($DE$) and false positive ratio ($FP$) to measure the accuracy of the recovered region undergone the same contrast enhancement transform. Specifically, with $R_D$ and $R_T$ corresponding to the detected and true region undergone $\phi_0$, respectively, with $|R|$ representing the area of an image region $R$, we define $DE = |R_T \cap R_D|/|R_T|, FP = 1 - |R_T \cap R_D|/|R_D|$. The average $DE$ and $FP$ of our method over all 100 images are 72.3% and 24.2%, respectively, with an averaged running time over a $2,000 \times 1,500$ image less than one minuteAs these results show, our method is capable of recovering the majority of the spliced regions.

On the other hand, we also noticed that large continuous regions with few pixel values due to large monotone or insufficient exposure can lead to false positive or mis-detections. The pixel histogram of these areas are usually sparse and a contrast enhancement transform just move bins around without changing the number of empty bins significantly. A future work is to identify such regions based on their pixel histogram and exclude them from the estimation procedure.

## 6 CONCLUSION

In this work, we describe a new method to estimate contrast enhancement from images, taking advantage of the nature of contrast enhancement as a mapping between integral pixel values and the distinct characteristics it introduces to the pixel histogramof the transformed image. Our method recovers the original pixel histogram and the applied contrast enhancement simultaneously with an efficient iterative algorithm, and can effectively handle perturbations due to noise and compression. We perform experimental evaluation to demonstrate the efficacy and efficiency of the proposed method. By examining local areas in the image, we also show that using this method, we can detect spliced image regions that have undergone different contrast enhancement transformations.

---

[7]We did not use the grid-search based Algorithm 4 to handle the parametric contrast enhancement as in practical scenarios we usually do not have the knowledge of the types of contrast enhancement involved.
[8]These images can be downloaded from https://mig.nist.gov/MediforBP/MediforBPResources.html.

There are several shortcomings to the current method which also provides directions we would like to further improve the current work in the future. First, our current algorithm is not very robust with regards to JPEG compression and we are investigating modeling JPEG compression using similar mathematical framework and handling them as in[8, 24]. Second, the current local estimation algorithm cannot effectively handle large image regions with monotone content or that are over or under exposure, and one idea is to automatically detect such areas and remove them from estimation. Third, our estimation method may not be able to handle contrast clipping, in which pixels of certain values are clipped. As such operation completely drops certain pixel values from and image, it may call for different estimation methods.

**Proof of Theorem 2.** First, for $x \in \mathcal{R}$, we have $|x| = \min_{z \geq 0} \frac{1}{2} \left( \frac{x^2}{z} + z \right) = \operatorname{argmin}_{z \geq 0} \frac{1}{2} \left( \frac{x^2}{z} + z \right)$, since for $x \neq 0$, differentiating the objective with regards to $z$ and setting results to zero give $\frac{1}{2} \left( -\frac{x^2}{z^2} + 1 \right) = 0 \Rightarrow z = |x|$, and for $x = 0$ the obvious optimal solution is $z = 0$. Using this result, we have

$$\|\mathbf{x}\|_1 = \sum_{i=1}^n |x_i| = \sum_{i=1}^n \min_{z_i \geq 0} \frac{1}{2} \left( \frac{x_i^2}{z_i} + z_i \right)$$
$$= \min_{z_i \geq 0} \sum_{i=1}^n \frac{1}{2} \left( \frac{x_i^2}{z_i} + z_i \right)$$
$$= \min_{\mathbf{z} \geq 0} \frac{1}{2} \left( \mathbf{x}^\top \mathcal{D}(\mathbf{z})^{-1} \mathbf{x} + \mathbf{1}^\top \mathbf{z} \right).$$

And so does the argmin part of the result.

**Projection on probability simplex.** The projection of a vector $\mathbf{x}$ on $\Delta^{n+1}$ is defined as the solution to the following optimization problem

$$\min_{\mathbf{h}} \frac{1}{2} \|\mathbf{x} - \mathbf{h}\|^2 \text{ s.t. } \mathbf{h} \geq 0, \mathbf{1}^\top \mathbf{h} = 1. \tag{26}$$

Introducing Lagrangian multipliers $\mathbf{y} \geq 0$ and $\xi$, we form the Lagrangian of Eq.(26) as

$$\mathcal{L}(\mathbf{h}, \mathbf{y}, \xi) = \frac{1}{2} \|\mathbf{x} - \mathbf{h}\|^2 - \mathbf{y}^\top \mathbf{h} - \xi(\mathbf{1}^\top \mathbf{h} - 1).$$

The corresponding KKT condition is then given by

$$\begin{aligned} \mathbf{h} - \mathbf{x} - \mathbf{y} - \xi \mathbf{1} &= 0 & & (\tfrac{\partial}{\partial \mathbf{h}} \mathcal{L}(\mathbf{h}, \mathbf{y}, \xi) = 0) \\ \mathbf{1}^\top \mathbf{h} &= 1 & & \text{(primal feasibility)} \\ 0 \leq \mathbf{y} &\perp \mathbf{h} \geq 0 & & \text{(complementary slackness)}. \end{aligned}$$

It is not hard to see that the following is a solution satisfying the KKT condition

$$\begin{aligned} \mathbf{y} &= (\mathbf{x} + \xi \mathbf{1})_+ - (\mathbf{x} + \xi \mathbf{1}) \\ \mathbf{h} &= (\mathbf{x} + \xi \mathbf{1})_+ \\ \xi &= \text{the solution of } \sum_{i=1}^n (x_i + \xi)_+ = 1. \end{aligned}$$

Here we define $(x)_+ = \max(x, 0)$ is the hinge function.

**Proof of Theorem 3.** To count the total number of different gamma correction transforms, we notice that if $i$ is mapped to $j$ by the gamma correction transform, we have

$$j - \frac{1}{2} \leq n \left( \frac{i}{n} \right)^\gamma < j + \frac{1}{2}.$$

This turns into $\underline{\gamma}_{ij} \leq \gamma < \bar{\gamma}_{ij}$ where

$$\underline{\gamma}_{ij} = \frac{\log n - \log(j + \frac{1}{2})}{\log n - \log i}, \bar{\gamma}_{ij} = \frac{\log n - \log(j - \frac{1}{2})}{\log n - \log i}.$$

We have (i) all $\underline{\gamma}_{ij}$ and $\bar{\gamma}_{ij}$ are distinct numbers and (ii) $\underline{\gamma}_{ij} = \bar{\gamma}_{i,j+1}$. As such, each different value of $\underline{\gamma}_{ij}$ signifies a change of gamma correction curves, while different values within the range $\underline{\gamma}_{ij} \leq \gamma < \bar{\gamma}_{ij}$ corresponds to the same curve. Therefore, the total number of gamma correction is bounded by $(n-1)^2$.

**Derivation of Algorithm 2.** Using Theorem 2, we introduce two auxiliary variables $\hat{\mathbf{u}} \succeq 0$ and $\hat{\mathbf{v}} \succeq 0$ to replace the $\ell_1$ norms in (17), and reformulate the problem as

$$\begin{aligned}
\min_{\hat{\mathbf{h}}, \hat{\mathbf{u}}, \hat{\mathbf{v}}} \quad & \hat{L}(\hat{\mathbf{h}}, \hat{\mathbf{u}}, \hat{\mathbf{v}}) \\
\text{s.t.} \quad & \mathbf{1}^T \hat{\mathbf{h}} = 1, \hat{\mathbf{h}} \succeq 0, \hat{\mathbf{u}} \succeq 0, \hat{\mathbf{v}} \succeq 0,
\end{aligned} \tag{27}$$

where

$$\begin{aligned}
\hat{L}(\hat{\mathbf{h}}, \hat{\mathbf{u}}, \hat{\mathbf{v}}) \quad &= \tfrac{1}{2}\mathbf{1}^\top \hat{\mathbf{u}} + \tfrac{\xi}{2}\mathbf{1}^\top \hat{\mathbf{v}} \\
&+ \tfrac{1}{2}\left(F\tilde{\mathbf{h}} - FR\hat{\mathbf{h}}\right)^\top \mathcal{D}(\hat{\mathbf{u}})^{-1}\left(F\tilde{\mathbf{h}} - FR\hat{\mathbf{h}}\right) \\
&+ \tfrac{\xi}{2}\left(F\hat{\mathbf{h}} - FT_\phi \mathbf{h}\right)^\top \mathcal{D}(\hat{\mathbf{v}})^{-1}\left(F\hat{\mathbf{h}} - FT_\phi \mathbf{h}\right).
\end{aligned}$$

Eq.(28) is a convex optimization problem jointly for $(\hat{\mathbf{h}}, \hat{\mathbf{u}}, \hat{\mathbf{v}})$, and we solve it also with a block coordinate descent scheme. Specifically, initializing $\hat{\mathbf{h}}_0$, $\hat{\mathbf{u}}_0$ and $\hat{\mathbf{v}}_0$, we find the optimal solution to it by iterating the following steps until convergence

- $\hat{\mathbf{h}}_{t+1} \leftarrow \text{argmax}_{\hat{\mathbf{h}}} \hat{L}(\hat{\mathbf{h}}, \hat{\mathbf{u}}_t, \hat{\mathbf{v}}_t);$
- $\hat{\mathbf{u}}_{t+1} \leftarrow \text{argmax}_{\hat{\mathbf{u}}} \hat{L}(\hat{\mathbf{h}}_{t+1}, \hat{\mathbf{u}}, \hat{\mathbf{v}}_t);$
- $\hat{\mathbf{v}}_{t+1} \leftarrow \text{argmax}_{\hat{\mathbf{v}}} \hat{L}(\hat{\mathbf{h}}_{t+1}, \hat{\mathbf{u}}_{t+1}, \hat{\mathbf{v}}).$

**Optimizing $\hat{\mathbf{h}}$**: fixing $\hat{\mathbf{u}}_t$ and $\hat{\mathbf{v}}_t$ and dropping irrelevant terms, minimizing $\hat{\mathbf{h}}$ reduces to the following constrained linear least squares problem

$$\begin{aligned}
\min_{\hat{\mathbf{h}}} \quad & \tfrac{1}{2}\hat{\mathbf{h}}^\top \hat{M}\hat{\mathbf{h}} - \hat{\mathbf{b}}^\top \hat{\mathbf{h}} \\
\text{s.t.} \quad & \mathbf{1}^T \hat{\mathbf{h}} = 1, \hat{\mathbf{h}} \succeq 0,
\end{aligned} \tag{28}$$

where

$$\begin{aligned}
\hat{M} &= (RF)^\top \mathcal{D}(\hat{\mathbf{u}}_t)^{-1} RF + \xi F^\top \mathcal{D}(\hat{\mathbf{v}}_t)^{-1} F \\
\hat{\mathbf{b}} &= (RF)^\top \mathcal{D}(\hat{\mathbf{u}}_t)^{-1} F\tilde{\mathbf{h}} + \xi F^\top \mathcal{D}(\hat{\mathbf{v}}_t)^{-1} FT_\phi \mathbf{h}.
\end{aligned}$$

Eq.(28) is solved with projected gradient descent: starting with $\mathbf{h}_{t,0} = \mathbf{h}_t$, our algorithm iterates between two steps:

$$\begin{aligned}
\triangle \hat{\mathbf{h}}_\tau &\leftarrow \hat{M}\hat{\mathbf{h}}_{t,\tau} - \hat{\mathbf{b}} \\
\hat{\mathbf{h}}_{t,\tau+1} &\leftarrow \mathcal{P}_{\Delta^{n+1}}\left(\hat{\mathbf{h}}_{t,\tau} - \tfrac{\eta_0}{\tau+1}\triangle \hat{\mathbf{h}}_\tau\right).
\end{aligned} \tag{29}$$

We take $\hat{\mathbf{h}}_{t+1} = \hat{\mathbf{h}}_{t,\tau}$ at the convergence.

**Optimizing $\hat{\mathbf{u}}$**: when fixing $\hat{\mathbf{h}}_{t+1}$ and dropping irrelevant terms, minimizing $\hat{\mathbf{u}}$ becomes

$$\begin{aligned}
\min_{\hat{\mathbf{u}}} \quad & \mathbf{c}^\top \mathcal{D}(\hat{\mathbf{u}})^{-1}\mathbf{c} + \mathbf{1}^\top \hat{\mathbf{u}} \\
\text{s.t.} \quad & \hat{\mathbf{u}} \succeq 0,
\end{aligned}$$

where $\mathbf{c} = F\left(\tilde{\mathbf{h}} - R\hat{\mathbf{h}}_{t+1}\right)$. Using Theorem 2, we obtain the optimal solution $\hat{\mathbf{u}}_{t+1} \leftarrow \left|F\left(\tilde{\mathbf{h}} - R\hat{\mathbf{h}}_{t+1}\right)\right|$.

**Optimizing $\hat{\mathbf{v}}$**: similarly, when fixing $\hat{\mathbf{h}}_{t+1}$ and dropping irrelevant terms, minimizing $\hat{\mathbf{v}}$ becomes

$$\begin{aligned}
\min_{\hat{\mathbf{v}}} \quad & \left(\hat{\mathbf{h}}_{t+1} - T_\phi \mathbf{h}\right)^\top F^\top \mathcal{D}(\hat{\mathbf{v}})^{-1} F\left(\hat{\mathbf{h}}_{t+1} - T_\phi \mathbf{h}\right) + \mathbf{1}^\top \hat{\mathbf{v}} \\
\text{s.t.} \quad & \hat{\mathbf{v}} \succeq 0.
\end{aligned}$$

Using Theorem 2 again, we have $\hat{\mathbf{v}}_{t+1} = \left|F\left(\hat{\mathbf{h}}_{t+1} - T_\phi \mathbf{h}\right)\right|$.